\documentclass[final]{arxiv}

\usepackage{times}
\usepackage{epsfig}
\usepackage{graphicx}
\usepackage{multirow}
\usepackage{amsmath}
\usepackage{amssymb}
\usepackage{breqn}
\usepackage{ctable}
\usepackage{booktabs}
\usepackage{xfrac}
\usepackage{afterpage}
\usepackage{bbm}

\usepackage{mathtools}

\renewcommand{\paragraph}[1]{\vspace{1mm}\noindent\textbf{#1}}
\newcommand{\Tref}[1]{Table~\ref{#1}}
\newcommand{\eref}[1]{Eq.~(\ref{#1})}

\newcommand{\fref}[1]{Fig.~\ref{#1}}

\newcommand{\sref}[1]{Sec.~\ref{#1}}

\usepackage[pagebackref=true,breaklinks=true,colorlinks,bookmarks=false]{hyperref}

\def\cvprPaperID{3674} 
\def\confYear{CVPR 2021}

\begin{document}

\title{Single-shot Path Integrated Panoptic Segmentation}

\author{Sukjun Hwang\\
Yonsei University\\
\and
Seoung Wug Oh\\
Adobe Research\\
\and
Seon Joo Kim\\
Yonsei University\\
}

\maketitle

\begin{abstract}
Panoptic segmentation, which is a novel task of unifying instance segmentation and semantic segmentation, has attracted a lot of attention lately.  
However, most of the previous methods are composed of multiple pathways with each pathway specialized to a designated segmentation task.
In this paper, we propose to resolve panoptic segmentation in single-shot by integrating the execution flows.
With the integrated pathway, a unified feature map called Panoptic-Feature is generated, which includes the information of both things and stuffs.
Panoptic-Feature becomes more sophisticated by auxiliary problems that guide to cluster pixels that belong to the same instance and differentiate between objects of different classes.
A collection of convolutional filters, where each filter represents either a thing or stuff, is applied to Panoptic-Feature at once, materializing the single-shot panoptic segmentation.
Taking the advantages of both top-down and bottom-up approaches, our method, named SPINet, enjoys high efficiency and accuracy on major panoptic segmentation benchmarks: COCO and Cityscapes.

\end{abstract}

\section{Introduction}

\begin{figure}
\centering
\includegraphics[width=1.0\linewidth]{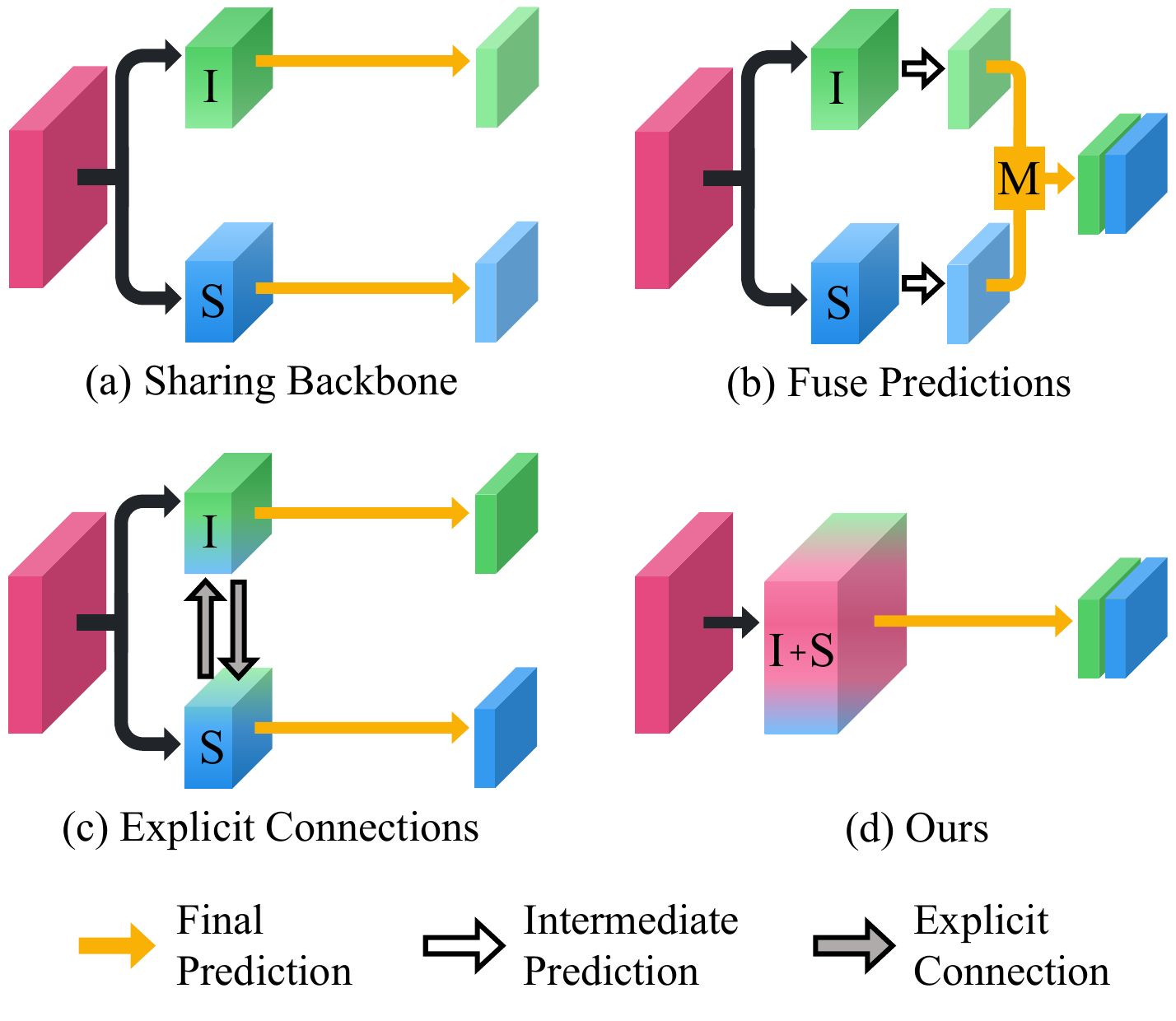}
\vspace{-2mm}
\caption{
{Panoptic segmentation with SPINet.}
(a) Most panoptic segmentation models tackle the task by separating the execution flow into two: instance branch $I$, and semantic branch $S$.
(b) Intermediate predictions from $I$ and $S$ can generate better final prediction by using an additional module $M$.
(c) Explicit connections between $I$ and $S$ alleviate the isolation of information.
(d) Our model integrates $I$ and $S$, performing panoptic segmentation within a single feature map. 
Through this integration, we can achieve improvement both in performance and efficiency.
}
\label{Fig:teaser}
\end{figure}

Panoptic segmentation, first proposed by Kirillov~\etal~\cite{PanopticSegmentation}, is a task with the goal of annotating each pixel to the corresponding category.
While semantic segmentation limits the term `category' to class labels, panoptic segmentation extends the term to include the concept of instances.
An early solution for the task was to use completely separate leading algorithms for instance and semantic segmentation~\cite{PanopticSegmentation}.
However, separating models for each task not only conflicts with the motivation behind panoptic segmentation but also doubles the computation and the model size.
As panoptic segmentation is based on the philosophy of unifying two different tasks, an ideal solution for the problem would be finding the best joint framework for both tasks.

To derive a joint solution for the new task, previous approaches started to adopt powerful instance or semantic segmentation models as baselines.
Most of currently leading panoptic segmentation models put their basis on either (top-down) Mask R-CNN~\cite{MaskRCNN} or (bottom-up) DeepLab~\cite{DeepLab}.
As the baseline itself cannot fully perform panoptic segmentation, additional modules become necessary: a semantic branch for Mask R-CNN~\cite{PanopticFPN} and an instance branch for DeepLab~\cite{DeeperLab}.

Efficiency-wise, the strategy of supplying additional modules on top of the chosen basis was successful~\cite{PanopticDeepLab, PanopticFPN, Seamless} (\text{\fref{Fig:teaser} (a)}).
Compared to executing two independent networks, the concept of sharing backbone layers for both instance and semantic branch led to a huge reduction of computations. 
Further improvement in performance was achieved by putting a connection between the two branches.
For instance, by including an extra module that takes intermediate predictions~\cite{Li_2020_CVPR, UPSNet} from the two branches resulted in performance improvement (\fref{Fig:teaser} (b)).
Moreover, supplying explicit connections between the branches~\cite{BANet, AttentionPanoptic, Bidirectional} as shown in \fref{Fig:teaser} (c), also enhanced the predictions as each branch complements from the information of their counterparts.
However, these methods still have separate pathways for instance and semantic segmentation, leaving room for further structural unification.



In this paper, we propose {\bf SPINet} with the goal of integrating the branches of instance and semantic segmentation (\fref{Fig:teaser} (d)).
By unifying two pathways, SPINet generates a single feature map called Panoptic-Feature, used for segmenting both thing and stuff.
In contrast to previous structures with separate branches, Panoptic-Feature holds the information of both thing and stuff, thus it bridges the gap between the two without any explicit connections.
Additionally, the computations necessary for capturing the context at high-level while retaining low-level fine details can be shared with the integrated pathway, making our model highly efficient. 

As we integrate the execution flow of thing and stuff altogether, the panoptic segmentation task can be finalized in \emph{single-shot} by applying a unified convolution to Panoptic-Feature.
The weights of thing classes are generated dynamically by reading the context of the input scene similar to CondInst~\cite{CondInst}, and trainable parameters are used for the weights of stuff classes.
By collecting all the weights and applying these to the Panoptic-Feature by the \emph{single-shot} convolution, SPINet can generate masks for every things and stuffs in the scene at once.

The key to the success of the proposed method lies in the representation power of the Panoptic-Feature.
To strengthen the Panoptic-Feature, we propose auxiliary tasks in order to learn a better latent space. 
The auxiliary tasks are designed to lead points in Panoptic-Feature to be clustered if they belong to the same class and moreover the same instance.
Trained with additional guidance signals from the auxiliary tasks, we show that the performance of our panoptic segmentation gains further improvement. 


SPINet achieves the state-of-the-art performance even without using heavy modules such as ASPP~\cite{DeepLab} or deformable convolution~\cite{DeformableConv}.
Moreover, by integrating the characteristics of both top-down and bottom-up, our model resolves the quality imbalance issue of previous approaches; top-down based models acquiring high PQ$^{th}$ but comparatively low PQ$^{st}$\footnotemark, and vice-versa.
For the first time, SPINet achieves comparable results to two-stage methods on COCO~\cite{COCO} without the region proposal network.
Moreover, our model delivers the state-of-the-art performance while yielding huge efficiency on Cityscapes~\cite{Cityscapes}.
With ResNet-50-FPN backbone, SPINet achieves 63.0\% PQ on Cityscapes \emph{val} set, and 42.2\% PQ on COCO \emph{val} set.

\afterpage{\footnotetext{PQ$^{th}$ and PQ$^{st}$ are \emph{panoptic quality} (PQ) averaged over \emph{thing} classes and \emph{stuff} classes respectively.~\cite{PanopticSegmentation}}}

\begin{figure*}[t]
\centering
\vspace{0.5mm}
\includegraphics[width=1.0\linewidth]{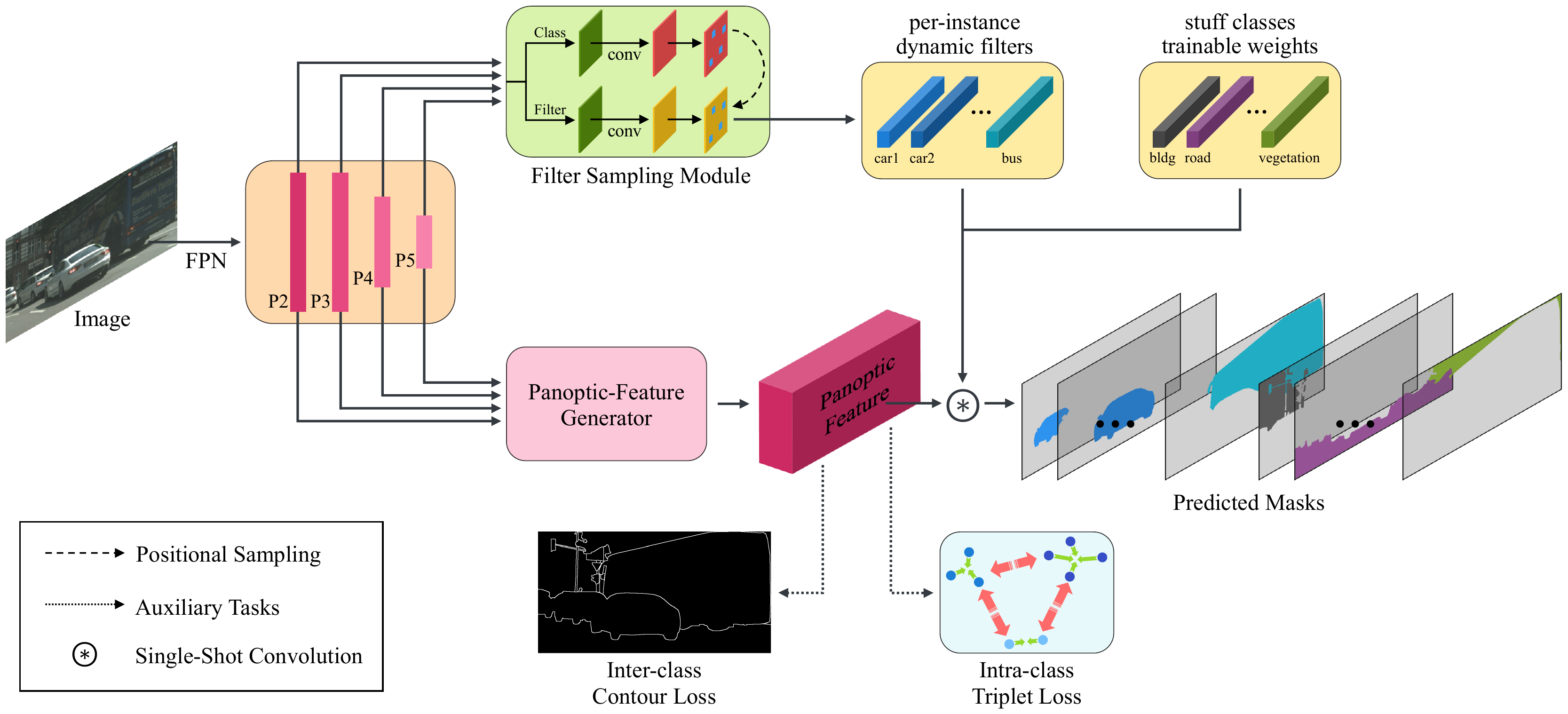}
\vspace{0.5mm}
\caption{
Overall execution flow of our proposed SPINet.
The architecture is mainly composed of three modules: FPN, filter sampling module and Panoptic-Feature generator (Fig.~\ref{fig:pan_feature}).
The three modules harmonize their use by the single-shot convolution, which generates masks for both instance and semantic segmentation at once.
}
\label{fig:main}
\end{figure*}

\section{Related Works}
\subsection{Panoptic Segmentation Approaches}
\paragraph{Mask R-CNN Based.}
Mask R-CNN~\cite{MaskRCNN} is a well-known instance segmentation model which detects instances first, then searches for finer details for segmentation.
Panoptic-FPN~\cite{PanopticFPN} proposed using Mask R-CNN for the new panoptic segmentation task with an additional semantic segmentation branch.
Numerous panoptic segmentation models~\cite{BANet, OCFusion, Li_2020_CVPR, AttentionPanoptic, EndtoEnd, Bidirectional, UPSNet} are built on top of Panoptic-FPN structure, increasing the performance with additional enhancement techniques.
Since `detect-then-segment' pipeline shows strength in capturing objects, panoptic segmentation methods based on  Mask R-CNN show high PQ$^{th}$.
However, as the models are not focused on preserving pixel-level details, PQ$^{st}$ is relatively low.

\paragraph{DeepLab Based.}
DeepLab~\cite{DeepLab} is a strong semantic segmentation model, which focuses on fine-grained details by taking encoder-decoder like structure to recover spatial resolution.
Panoptic segmentation models that adopt DeepLab ~\cite{PanopticDeepLab, DeeperLab} not only generate pixel-level predictions for stuff but also for thing.
With the fine-grained predictions, formation of instances is possible by a bottom-up methodology; aggregating the pixels that have similar aspects.
As opposed to top-down models with Mask R-CNN, bottom-up panoptic segmentation models acquire high PQ$^{st}$, but low PQ$^{th}$.

\paragraph{DETR.}
Unifying thing and stuff classes altogether, DETR~\cite{DETR} predicts which categories exist in the input.
With a number of attention heatmaps, generated per each predicted category by the transformer~\cite{Transformer}, mask logits corresponding to each category can be generated.
The final prediction of panoptic segmentation can be done by a single pixel-wise \texttt{argmax} on the predicted mask logits.
Though the methodology is noticeably simple, DETR requires an extremely large computing resource as it needs to decode the attention heatmaps for every instances separately in addition to the heavy base computation from the use of the transformer.

SPINet shows aspects of \emph{bottom-up} as each point of Panoptic-Feature retains information about which class and instance they originate from.
Meanwhile, the generation of dynamic filters catches the context of instances~\cite{CondInst}, hence our model can also be viewed as \emph{top-down}.
Similar to DETR, SPINet integrates the execution flow of thing and stuff by treating the both equally, but lighter as SPINet requires less steps to generate predictions. Taking the advantages of top-down, bottom-up, and unified execution flow, SPINet resolves the performance imbalance issue, showing competitive results on both COCO and Cityscapes with efficiency.

\subsection{One-stage Instance Segmentation}
Latest object detection and instance segmentation models can be categorized by the existence of region proposal network (RPN)~\cite{FasterRCNN}: one-stage~\cite{RetinaNet, SSD, YOLO} and two-stage~\cite{CascadeRCNN, MaskRCNN, FasterRCNN}.
After the emergence of Mask-RCNN~\cite{MaskRCNN}, the majority of instance segmentation models is built on top of two-stage detectors, hence improvement in object detection spontaneously led to performance gain in instance segmentation.
RPN shows substantial performance, yet the extraction of regional features remains as a bottleneck.
To overcome the issue, there has been a lot of improvement on one-stage detectors such as YOLO~\cite{YOLO}, and  FCOS~\cite{FCOS}. 
Many one-stage methods for instance segmentation~\cite{CondInst, SOLOv1, CenterMask} are now showing comparable performance to that of two-stage based methods.

Recently proposed CondInst~\cite{CondInst} generates dynamic convolutional weights that correspond to an instance for each feature location.
Our model adopts dynamic filters proposed by CondInst for segmenting instances.
By equally treating the filters for thing and stuff, SPINet can finalize the panoptic segmentation with a single-shot convolution.

\section{SPINet}
\label{sec:SPINet}
As the primary goal of SPINet is to integrate the pathways of thing and stuff, our model has a simple execution flow as illustrated in~\fref{fig:main}.
SPINet is composed of four steps: FPN backbone, filter sampling module, Panoptic-Feature generator, and single-shot convolution.
First, the input image is encoded into multi-scale features through FPN backbone.
Then, Panoptic-Feature generator takes the multi-scale features and constructs Panoptic-Feature.
At the same time, the filter sampling module also operates on each stage of the multi-scale features to dynamically generate filters for segmenting instances in the scene.
Filters for semantic segmentation are defined as trainable parameters and learned through back-propagation. 
Finally, single-shot convolution with dynamic filters for things and learned filters for stuffs are applied to Panoptic-Feature to finalize the overall prediction at once.
Further improvement of performance is possible by adding auxiliary tasks that enhance the representation power of the Panoptic-Feature.

\subsection{FPN}
\label{sec:fpn}
We utilize FPN~\cite{FPN} as multi-level features of our model.
A considerable number of detection models adopting FPN use five levels of the feature maps $\{P_3, P_4, P_5, P_6, P_7\}$.
However, using these levels results in favoring instance segmentation over semantic segmentation.
To deal with this issue, we make a slight adjustment, employing relatively low-level feature maps of $\{P_2, P_3, P_4, P_5\}$.
Since a feature map of stride 4 has substantial spatial size, passing it to convolutional layers results in excessive computations.
Therefore, we scale down $C_2$ in half, generating $P_2$ with a summation between halved $C_2$ and $P_3$.
Note that for COCO, we include additional level $P_6$, and the generation of $P_6$ is the same as $P_5$, which is of stride 32.
\subsection{Filter Sampling Module}
We adopt dynamic convolutional filters, inspired by CondInst~\cite{CondInst}, to design our filter sampling module that is composed of the class head and the filter head.
To deal with objects of different scales, the filter sampling module takes multi-level features $P_l$ from FPN as inputs.
The weights of the module are shared across all levels.
With the class and the filter head, the filter sampling module predicts the class scores and generates dynamic convolutional filters, each corresponding to a designated instance.
The class head consists of four sequential convolutions which outputs $C_l \in \mathbb{R}^{H_l \times W_l \times N_t}$, where $H_l$ and $W_l$ are the height and width of the feature $P_l$, and $N_t$ is the number of thing classes (e.g. 8 for Cityscapes dataset).
For the filter head, the structure is the same as the class head, but it additionally takes the absolute positional information encoded at each level~\cite{CoordConv}.
The filter head produces dynamic filter $F_l \in \mathbb{R}^{H_l \times W_l \times D_f}$ as an output, where $D_f$ is the number of output channels.


Let $k$ be the size of the kernel used for the single-shot convolution.
A filter of ${k^2}D_f$ dimensions can be obtained from the output of the filter head $F_l$; spatially pooled by the size of $k \times k$, putting the center to a given sampling location.
During the training, the filters are uniformly sampled from the foreground locations.
On the other hand, for the inference, the positions to be sampled are selected by the confidence score above a threshold, measured by $C$.
The sampled filters can be finally used as convolutional weights after passing through a fully connected layer, each resulting in ${k^2}D_\phi$ of the number of channels, where $D_\phi$ is the number of channels of the Panoptic-Feature.
The separation between the convolution and the fully connected layer can save memory and computation by forwarding only the filters at interested locations.

\subsection{Panoptic-Feature Generator}
\begin{figure}
\centering
\includegraphics[width=1.0\linewidth]{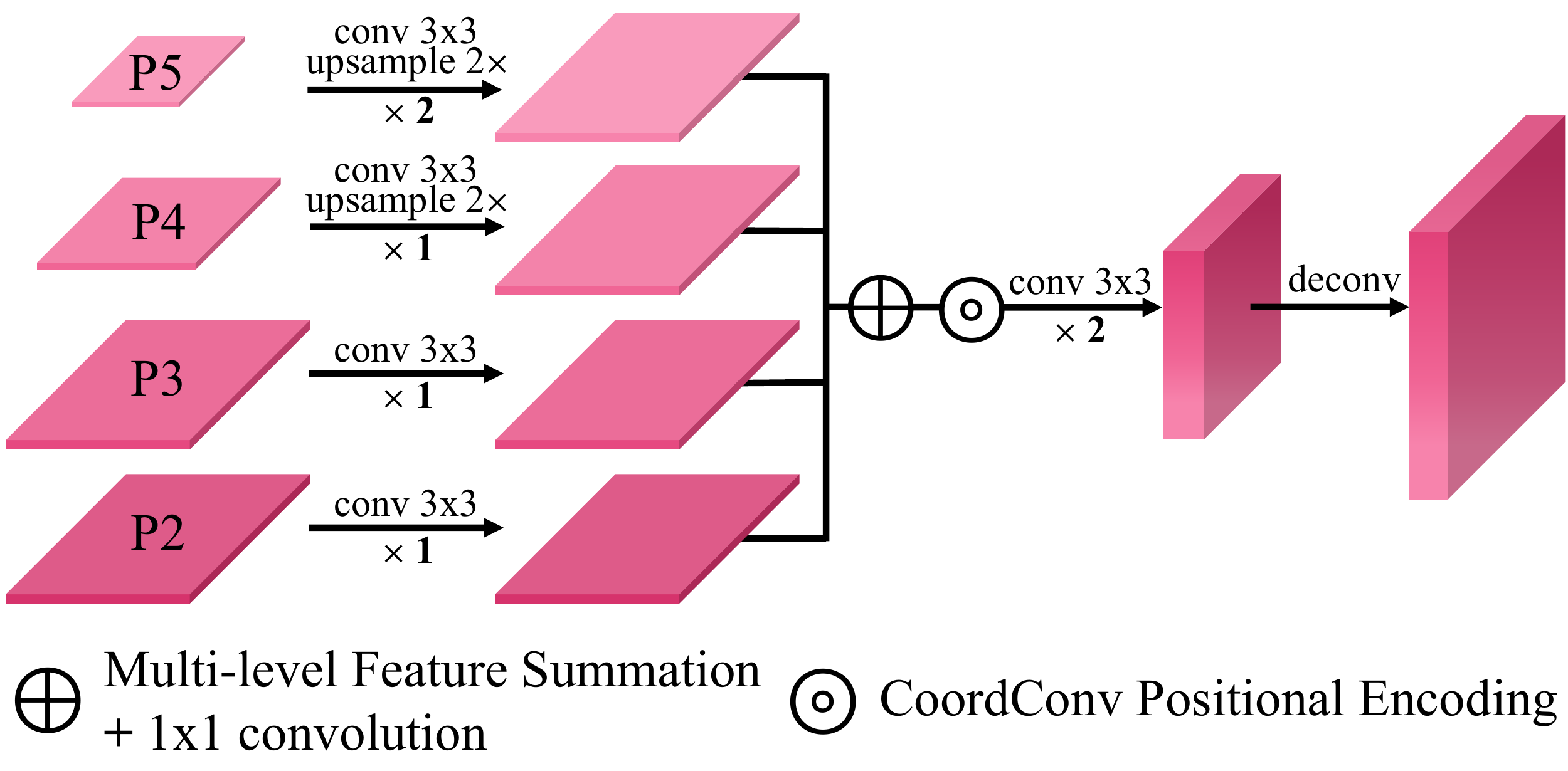}
\caption{
Panoptic-Feature generator.
By merging multi-level features from $P_2$ to $P_5$, the feature map holds information from low-level to high-level altogether.
Given the positional encoding, it passes through extra layers and become the size of stride 4.
}
\vspace{-3mm}
\label{fig:pan_feature}
\end{figure}

The structure of Panoptic-Feature generator is similar to the segmentation branch of Panoptic-FPN~\cite{PanopticFPN}.
As the Panoptic-Feature is generated with the multi-level information from FPN, the features can simultaneously capture the context while fine-grained details are preserved.
Note that per-instance relative coordinates can provide strong cues for segmenting instances as proposed in CondInst~\cite{CondInst}.
However, since all things and stuffs should be handled together in the Panoptic-Feature, per-instance relative coordinates cannot be used.
Rather, as shown in~\fref{fig:pan_feature}, Panoptic-Feature generator utilizes the absolute positional encoding of CoordConv~\cite{CoordConv} to the merged output of the multi-level features, passing two additional convolutions followed by a deconvolutional layer.

To make the output features to have access to information of distant spatial locations, many previous works supply additional modules to their segmentation branch.
The use of convolutions with deformation~\cite{DeformableConv}, dilation~\cite{DeepLabv3plus}, or concatenating features from different levels have proven to be powerful~\cite{Li_2020_CVPR, Seamless, UPSNet}.
Compared to general semantic segmentation branches that adapt each level to the size of stride 4, Panoptic-Feature generator internally maintains its spatial size to the stride of 8.
As the size of the Panoptic-Feature is more compact, it can reach the information of distant locations by passing through convolutional layers.
Therefore, SPINet can be accurate and efficient as it avoids using the aforementioned heavy computations.
Moreover, by training with our proposed auxiliary tasks, the representation of the Panoptic-Feature can be further enriched.
\subsection{Unified Single-shot Convolution}

In our panoptic segmentation framework, the segmentation of both things and stuffs can be done by a single-shot convolution.
The convolutional weights for things are sampled from the filter sampling module and the weights for stuffs are learned as trainable parameters.
Applying these weights together on the Panoptic-Feature, the raw mask logits responsible for things $\hat{y}_t$ and stuffs $\hat{y}_s$ are generated.

We have separate losses for things and stuffs, denoted as $\mathcal{L}_t$ and $\mathcal{L}_s$.
These losses are computed using the ground truth for segmentation ($y_t$ and $y_s$).
For the loss of things $\mathcal{L}_{t}$, we uniformly sample $y_t$ and $\hat{y}_t$ from foreground locations, and scaled to the stride of 4.
The loss is computed as follows:
\begin{gather}
    \mathcal{L}_{t}(y_t, \hat{y_t}) = \textsc{Dice}(y_t, \sigma(\hat{y}_t)),
    \label{eqn:thing}
\end{gather}
where $\sigma$ is the sigmoid function, and \textsc{Dice} is the dice loss as in~\cite{VNet}.
$\mathcal{L}_{t}$ is normalized by the number of sampled sets.

For the loss of stuffs $\mathcal{L}_{s}$, we use a summation of two losses: bootstrapped cross entropy loss ($\mathcal{L}_{ce}$)~\cite{BootstrapCE} and the new multi-class dice loss ($\mathcal{L}_{md}$).
We refer readers to~\cite{BootstrapCE} for the details of $\mathcal{L}_{ce}$. We make use of the dice loss again, and $\mathcal{L}_{md}$ is computed as follows:
\begin{equation}
    \label{eqn:multi-dice}
    \mathcal{L}_{md}(y_s, \hat{y_s}) = \textsc{Dice}(y_s, \psi(\hat{y}_s)),
\end{equation}
where softmax -- taken across stuff classes -- is used for $\psi$, and $\mathcal{L}_{md}$ is normalized by the number of sampled sets, which in this case equals to the number of stuff classes.

The difference between \eref{eqn:thing} and \eref{eqn:multi-dice} is the way it ranges the logits between 0 and 1 (e.g. sigmoid and softmax).
Note that a finer semantic segmentation is shown possible with the simple use of Semantic Encoding Loss~\cite{EncNet}, which guides to better predict the presence of each class in the input.
Similar to Semantic Encoding Loss, multi-class dice loss guides the mask of each class to be better fitted, enabling finer segmentation results.

\begin{figure}
\centering
\includegraphics[width=1.0\linewidth]{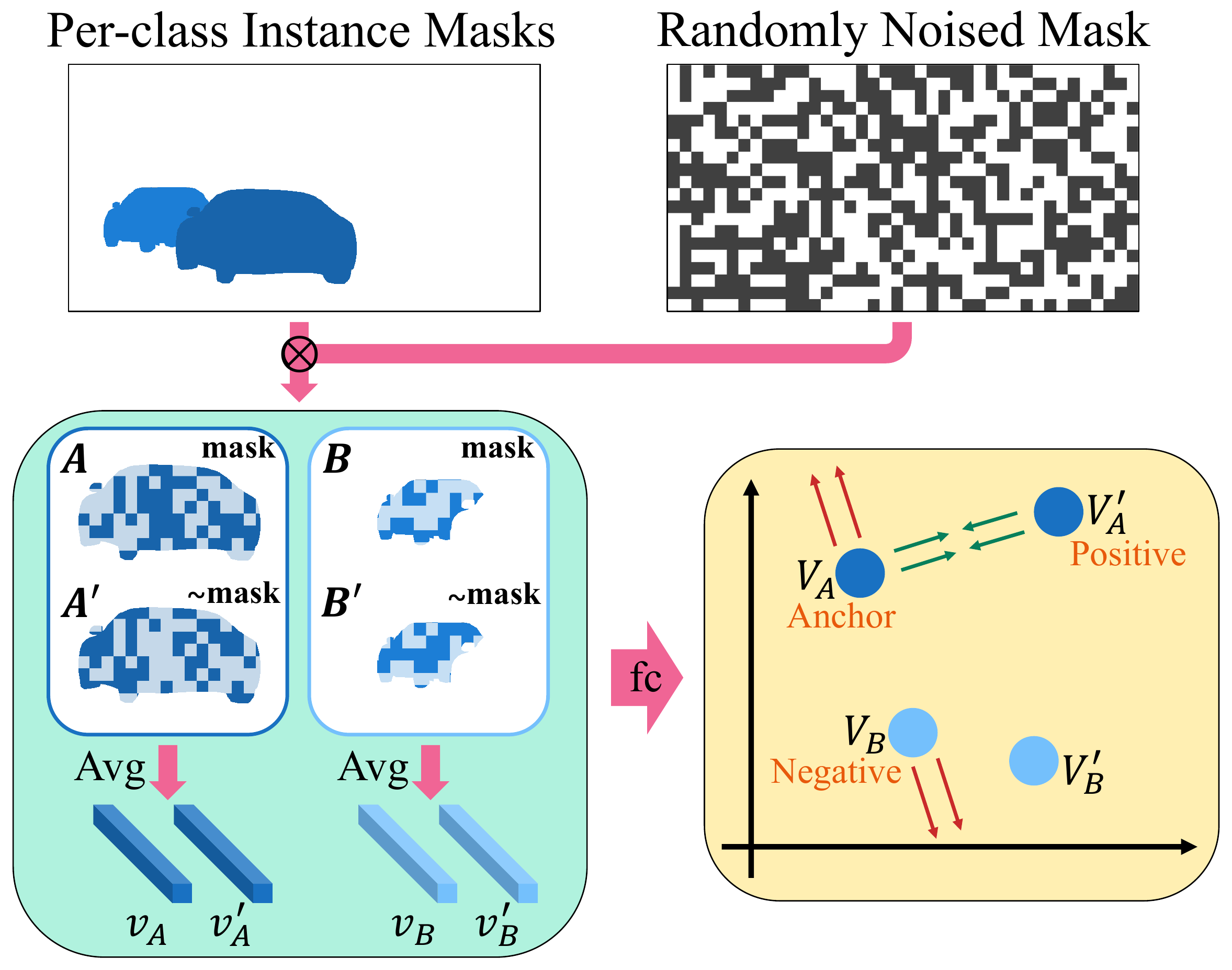}
\caption{
Intra-Class Triplet loss.
A randomly noised mask splits an instance mask to a pair.
With multiple representations of instances, distances between each other at latent space can be measured.
}
\vspace{-2mm}
\label{fig:triplet}
\end{figure}
\section{Auxiliary Tasks for Clustering}
\label{sec:auxiliary}
One way to interpret the Panoptic-Feature is as follows: each point in Panoptic-Feature clusters together at latent space if they belong to a same class and moreover same instance.
From this interpretation, Panoptic-Feature can be viewed as a huge composition of groups.
In the latent space, there exists a large gap between the clusters of different classes, whereas the gap between instances of a same class is comparatively narrow.

Recently, many approaches improved performance by making use of auxiliary tasks that are designed properly with the consideration to the main task~\cite{Yolact, Wu_2020_CVPR}.
We also offer new auxiliary tasks considering the interpretation of the Panoptic-Feature given above. 
The tasks are designed to give better guidance to the Panoptic-Feature.
For example, the distance between points in the Panoptic-Feature increases if they are from different instances, and vice-versa.
Given the guidance, the latent space of the Panoptic-Feature arranges properly, allowing the single-shot convolution to make better predictions.

\paragraph{Intra-Class Triplet Loss.}
Let $\phi \in \mathbb{R}^{H \times W \times D_\phi}$ be the Panoptic-Feature with $D_\phi$ number of channels over its $H \times W$ spatial size.
Let there exists an instance $k$ with the ground-truth mask $m_k \in \{0,1\}^{H \times W}$.
Given a randomly noised mask $m_r \in \{0,1\}^{H \times W}$, we can obtain a pair of partial masks $M_k, M_k^\prime$, both corresponding to the instance $k$ as follows:
\begin{equation}
    M_k = m_k \odot m_r,{\qquad}M_k^\prime = m_k \odot ({\sim}m_r),
\end{equation}
where ($\sim$) operator inverses the value of each pixel, and $\odot$ is the element-wise multiplication.

With a partial mask $M$, we can generate a representation vector $V$ as follows:
\begin{equation}
\begin{gathered}
    v^c = \frac{1}{N_M} \sum_{i,j} M_{i,j} \ast \phi_{i,j}^c,{\quad}v = \left[v^1, v^2,...,v^{D_\phi}\right]\\
    V = \textsc{FC}(v),
\end{gathered}
\end{equation}
where $M_{i,j}$ is the mask value of location $(i,j)$, and $N_M$ is the number of locations where $M_{i,j}=1$.
$\phi_{i,j}^c$ is the value of $c^{th}$ channel at location $(i,j)$ of the Panoptic-Feature, and \textsc{FC} is a fully connected layer.
By taking the above strategy, a representation pair $V_k$ and $V_k^\prime$ can be obtained from the pair of partial masks $M_k$ and $M_k^\prime$.

A collection of representation pairs from instances can be gathered by applying the same to each instance.
From the collection, triplets can be obtained by assigning representations from a same instance to pair up into $Anchor$ and $Positive$, while a representation from another instance become $Negative$.
For tasks like re-identification and metric learning, off-the-shelf triplet margin loss~\cite{FaceNet} is a great choice to be considered.
In order to better cluster points, we use Intra-Class Triplet Loss for narrowing the point features belonging to a same instance, and spreading the clusters of different instances.
After some modifications from the triplet margin loss, our loss is defined as follows:
\begin{equation}
\begin{gathered}
    d_p = - \textsc{Dist}(V_a, V_p),{\qquad}d_n = - \textsc{Dist}(V_a, V_n),\\
    \mathcal{L}_{intra}(V_a, V_p, V_n) = -\log\left(\frac{e^{d_p}}{e^{d_p} + e^{d_n}}\right),
\end{gathered}
\end{equation}
where $\mathcal{L}_{intra}$ is normalized by the number of triplet sets.
$\textsc{Dist}$ measures the L2 distance of an input pair, and $V_a$, $V_p$, $V_n$ are representation vectors of $Anchor$, $Positive$, and $Negative$ respectively.
An illustration of the triplet loss is given in~\fref{fig:triplet}.

\paragraph{Inter-Class Contour Loss.}
We assign an additional task to the Panoptic-Feature which is of predicting contours between different classes.
An $1\times1$ convolutional layer is applied to the Panoptic-Feature, and the output channel of the layer is 16.
The spatial size of the output is the same as the Panoptic-Feature, which is of stride 4.
Hence each location takes charge of $4\times4$ area with the pixel-shuffle~\cite{PixelShuffle} operation on the 16 channels.
Inter-Class Contour Loss ($\mathcal{L}_{inter}$) is computed over every pixel as follows:
\begin{gather}
    \mathcal{L}_{inter} = \frac{1}{{H}\times{W}}\sum_{i=1}^{H}\sum_{j=1}^{W} \textsc{Focal}(y_{i,j}, \hat{y}_{i,j}),
\end{gather}
where $\textsc{Focal}$ is the focal loss~\cite{RetinaNet}.
$y_{i,j}$ and $\hat{y}_{i,j}$ are the value of ground truth and predicted soft contour at location $(i, j)$, respectively.
The task of predicting contours from the Panoptic-Feature leads to enlargement of distances between class-wise clusters in latent space, and furthermore it eases the convolutional weights to distinguish classes.

Trained with the introduced Intra-Class Triplet Loss and Inter-Class Contour Loss, our model differentiates points from contrasting classes, and also clusters by instances.
The auxiliary tasks stabilize our model's training, and SPINet improves performance while not compromising its efficiency as the tasks are used only for training.

\begin{table*}
\centering
\begin{tabular}{cl|cc|ccccc|ccccc}
\toprule
\multicolumn{4}{c}{} & \multicolumn{5}{c}{COCO \emph{val} set} & \multicolumn{5}{c}{Cityscapes \emph{val} set}\\
\midrule
& Method & \emph{deform} & \emph{atrous} & PQ & PQ$^{th}$ & PQ$^{st}$ & AP & mIoU & PQ & PQ$^{th}$ & PQ$^{st}$ & AP & mIoU\\
\midrule
\midrule
\multirow{5}{*}[-0.4ex]{\rotatebox{90}{w/  RPN}}
& Pan-FPN~\cite{PanopticFPN} & & & 39.0 & 45.9 & 28.7 & 33.3 & 41.0 & 57.7 & 51.6 & 62.2 & 32.0 & 75.0\\
& OCFusion~\cite{OCFusion} & \checkmark & & 42.5 & 49.1 & 32.5 & - & - & 59.3 & 53.5 & 63.6 & - & -\\
& UPSNet~\cite{UPSNet} & \checkmark & & 42.5 & 48.6 & 33.4 & 34.3 & \underline{54.3} & 59.3 & 54.6 & 62.7 & 33.3 & 75.2\\
& Seamless~\cite{Seamless} & & \checkmark & - & - & - & - & - & 60.3 & 56.1 & 63.3 & 33.6 & 77.5\\
& Li~\etal\cite{Li_2020_CVPR} & \checkmark & & \underline{43.4} & 48.6 & \underline{35.5} & \underline{36.4} & 53.7 & 61.4 & 54.7 & 66.3 & 33.7 & 79.5\\
\midrule
\multirow{4}{*}[-0.4ex]{\rotatebox{90}{w/o RPN}}
& DeeperLab$^\dagger$~\cite{DeeperLab} & & \checkmark & 33.8 & - & - & - & - & 56.6 & - & - & - & -\\
& SSAP$^\ddagger$~\cite{SSAP} & & & 36.5 & - & - & - & - & 58.4 & 50.6 & - & 34.4 & - \\
& Hou~\etal~\cite{RealTimePanoptic} & & & 37.1 & 41.0 & 31.3 & - & - & 58.8 & 52.1 & 63.7 & 29.8 & 77.0\\
& Pan-DL$^\star$~\cite{PanopticDeepLab} & & \checkmark & 35.1 & - & - & - & - & 60.3 & 51.1 & 67.0 & 33.2 & 78.2\\
\midrule\midrule
& SPINet & & & \bf{42.2} & \bf{\underline{49.3}} & \bf{31.4} & \bf{33.2} & \bf{43.2} & \bf{\underline{63.0}} & \bf{\underline{57.0}} & \bf{\underline{67.3}} & \bf{\underline{35.3}} & \bf{\underline{80.0}}\\
\bottomrule
\end{tabular}
\caption{
Performance on COCO \& Cityscapes validation sets.
We underline the highest numbers among all, and bold the highest numbers over the models without RPN.
For fairness, the scores are measured without flipping and multiscale input evaluation, and the models are trained on top of ResNet-50, using ImageNet pretrained weights.
\emph{deform}: Deformable convolution is used.
\emph{atrous}: Atrous convolution is used.
$\dagger$: Xception-71 is used for backbone.
$\ddagger$: ResNet-101 is used for backbone with horizontal flipping and mutliscale test.
$\star$: Scores of Cityscapes are from latest \texttt{Detectron2}~\cite{Detectron2}.
}
\vspace{-3mm}
\label{table:comparison}
\end{table*}
\begin{table}
\centering
{\footnotesize
\begin{tabular}{l|c|c|ccc}
\toprule
\multirow{2}{*}{Method} & \multirow{2}{*}{PQ (\%)} & \multirow{2}{*}{Params (M)} & \multicolumn{3}{c}{Inference Time (ms)}\\
& & & Total & Network & Post\\
\midrule
\midrule
UPSNet & 59.3 & 44.5 & 501 & 191 & 310\\
Pan-DL & 60.3 & 59.8 & 499 & 335 & 164\\
\midrule
SPINet & 63.0 & 42.2 & 201 & 171 & 30\\
\bottomrule
\end{tabular}
}
\caption[Caption for LOF]{
Inference time on Cityscapes \emph{val} set.
We used codes that are published by the authors.
{\bf Total:} Network + Post. {\bf Network:} Time for a model to complete feed-forward execution. {\bf Post:} Time for post-processing.
}
\vspace{-3mm}
\label{Table:speed}
\end{table}
\section{Experiments}
\label{sec:experiments}
In this section, we compare our results with previous works on standard panoptic segmentation benchmarks.
We analyze our model with the models that are on par with the state-of-the-art performance.
We also provide ablation studies conducted on various settings.
All experiments are conducted using ResNet-50 as the backbone.
\subsection{Datasets}
Our experiments are conducted on two popular benchmark datasets: COCO and Cityscapes.
{\bf COCO}~\cite{COCO} is a large dataset that has annotations of 118K, 5K, and 20K images for training, validation, and testing, respectively.
The number of thing classes is 80, and stuff classes is 53.
We use fine-grained annotations of {\bf Cityscapes}~\cite{Cityscapes} which is composed of 2975 images for training and 500 for validation.
The dataset has 8 thing classes and 11 stuff classes, and is considered as a great benchmark for panoptic segmentation task due to its highly delicate annotations.

\subsection{Implementation Details}
\label{sec:imp_details}
Unless specified, we use ImageNet pretrained ResNet-50 backbone, that replaces the first $7\times7$ convolution with three $3\times3$ convolutions, which has proven to be effective by He~\etal~\cite{ResNetStem}.
We discuss the effect of this modification in~\sref{sect:ablation}.
All convolutional layers of filter sampling module are followed by group normalization~\cite{GroupNorm}, and we freeze batch normalization of the layers of the backbone.
The code we use will be made available.

\paragraph{Training.}
With all the losses introduced, we can now finalize the total loss of SPINet as follows:
\begin{equation}
    \mathcal{L} = \lambda_0\mathcal{L}_{cls} + \lambda_1\mathcal{L}_{s} + \lambda_2\mathcal{L}_{t} + \lambda_3\mathcal{L}_{inter} + \lambda_4\mathcal{L}_{intra},
\end{equation}
where $\mathcal{L}_{cls}$ uses focal loss~\cite{RetinaNet}.
For $\mathcal{L}_{ce}$ of $\mathcal{L}_{s}$, bootstrapped cross entropy loss of 0.2 top-k ratio is used for Cityscapes, while COCO uses general cross entropy.
The coefficients $(\lambda_0, \lambda_1, \lambda_2, \lambda_3, \lambda_4)$ differs by the dataset.
Cityscapes uses the coefficients of (1, 1, 5, 20, 1), while COCO adopts (1, 0.5, 3, 0, 1).
Our model can benefit from training with $\mathcal{L}_{inter}$ on Cityscapes as the dataset is composed of fine-grained pixel-level annotations.
However, the coarse annotations of COCO diminishes the enhancement.

Using 8 Tesla V100 GPUs, our models are trained for 270k iterations, allocating 2 images per GPU on COCO experiments.
The learning rate starts from 0.01 and decreases by a scale factor of 0.1 at 180k and 240k.
Allocating 4 images per GPU for Cityscapes experiments, the learning rate starts from 0.01 and decreases at 80k and 90k, terminating the training at 95k.

\paragraph{Inference.}
Most of the forward procedures are as elaborated in~\sref{sec:SPINet}, and we preserve the full resolution of images from all datasets.
We forward the input image through the network and give the confidence threshold of 0.45 for the sampling of dynamic filters for instances.
Additional steps for the post-processing follows that of Panoptic-FPN~\cite{PanopticFPN}, which marks down instances first, and adds up the predictions of stuff classes to the non-instance regions.

\subsection{Comparison to the State of the Art}
\label{sec:comparison}
To highlight our model's performance, we compare our proposed SPINet to the state-of-the-art models for both COCO and Cityscapes dataset as shown in~\Tref{table:comparison}.
For fair comparison, we listed performances of each models that are built on top of ResNet-50 and pretrained on ImageNet dataset.

COCO and Cityscapes have contrastive charateristic to each other.
COCO is composed of numerous instances, thus the importance lies in the ability to capture instances.
Meanwhile, to achieve high scores in Cityscapes, the ability to precisely predict the class of pixels is important.
Hence, there has been a tendency of RPN based models being dominant on COCO, and non-RPN based models favoring Cityscapes until now.
This tendency is clearly shown in~\Tref{table:comparison}, where previous models without RPN suffer from low performance on COCO dataset while being competitive on Cityscapes.
However, with the strong ability of capturing instances, our model's performance on COCO is substantially higher than all non-RPN based model.
Furthermore, the maintenance of pixel level details allows SPINet to surpass all previous models on Cityscapes, becoming the new state-of-the-art.

We also compare our model's inference speed with two well-known models: UPSNet~\cite{UPSNet} and Panoptic-DeepLab~\cite{PanopticDeepLab} where each represents a model with and without RPN.
We measured time to output raw predictions, and time to post-process raw predictions.
All times are measured on the same system environment, using TITAN XP.
Note that Panoptic-DeepLab gains more accuracy than the proposed results in the paper by using heavier heads, and modifying the first convolutional layer of the backbone as described in~\sref{sec:imp_details}~\cite{Detectron2}.
Our model enjoys the integrated structural pathway, where it omits both feature extraction at RPN and using heavy modules as ASPP.
SPINet accelerates its prediction while performing the best score as shown in ~\Tref{Table:speed}.

\begin{figure*}
\begin{center}
\includegraphics[width=1.0\linewidth]{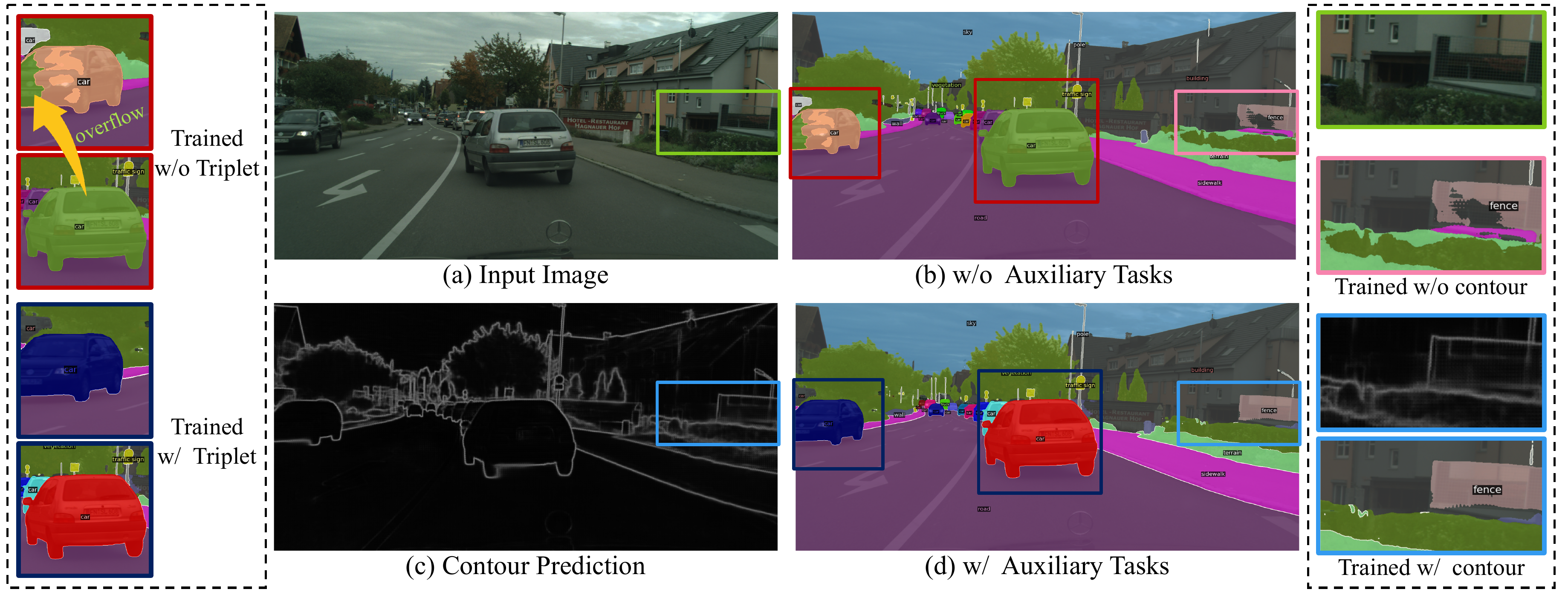}
\end{center}
\vspace{-4mm}
\caption{
Visualized results of SPINet with ResNet-50 on Cityscapes \emph{val} set.
With the given image (a), we show two panoptic segmentation results, which are retrieved from the models trained without auxiliary tasks (b), and with auxiliary tasks (d).
As the model trained with auxiliary tasks (d) uses two additional losses, Intra-Class Triplet loss and Inter-Class Contour loss, we visualize how the losses affect the prediction.
We also visualized the contour map (c), predicted by (d), to ease the understanding of how the contours can increase the segmentation quality.
}
\label{fig:visualization}
\vspace{-4mm}
\end{figure*}

\subsection{Ablation Study}\label{sect:ablation}
We conduct ablation studies on COCO and Cityscapes validation sets using our model with ResNet-50.
We show how our model can be improved by differentiating the model under various settings.

\paragraph{Use of Proposed Losses.}
As shown in ~\Tref{table:aux}, the performance of SPINet can be improved by making use of the proposed losses.
Note that the modules needed for the auxiliary tasks are not used during inference, thus the use of auxiliary tasks does not burden the model.
Using multi-class dice loss and both auxiliary tasks enhances the segmentation quality, and also stabilizes the training.
By taking a look at \fref{fig:visualization}, we can understand how the auxiliary tasks affect the predictions.
Compared to the model trained without auxiliary tasks, the model trained with auxiliary tasks tries to avoid the overflow of segmentation masks as Intra-Class Triplet loss restricts the intervention between instances as shown in \fref{fig:visualization} (b) and (d).
Furthermore, SPINet trained with Inter-Class Contour loss detects the borders between semantic classes.
Therefore, with the potential to estimate contours, our model can generate more accurate segmentation maps.

\begin{table}
\centering
{\footnotesize
\begin{tabular}{cc|c|c|c|c}
\toprule
M.C.D & Aux & PQ (\%) & PQ$^{th}$ (\%) & PQ$^{st}$ (\%) & Params (M)\\
\midrule\midrule
& & 61.8 & 54.8 & 66.8 & 42.2 \\
\checkmark & & 62.3 & 56.0 & 66.8 & 42.2 \\
& \checkmark & 62.3 & 55.9 & 66.9 & 42.3 \\
\checkmark & \checkmark & {\bf 63.0} & {\bf 57.0} & {\bf 67.3} & 42.3 \\
\bottomrule
\end{tabular}
}
\caption{
Use of different losses on Cityscapes \emph{val} set.
We highlight the effectiveness of our proposed losses.
{\bf M.C.D}: Multi-class dice loss. {\bf Aux}: Using both auxiliary tasks.
{\bf Params}: Total number of parameters needed for training.
}
\vspace{-3mm}
\label{table:aux}
\end{table}

\paragraph{Different FPN Level.}
Since our model uses the single Panoptic-Feature, balancing the performance between thing and stuff becomes important.
SPINet uses the multi-level features from FPN in two modules: Panoptic-Feature generator and filter sampling module.
From experiments, we find that using the lower levels from FPN results in the best performance as shown in \Tref{Table:FPNlevel}.
Therefore, we lower the resolution of P2 to balance the performance and computation loads as described in~\sref{sec:fpn}, and use the downscaled feature for both Panoptic-Feature generator and filter sampling module.

\begin{table}
\centering 
\begin{tabular}{c|c|ccc}
\toprule
P.F.G & F.S.M & PQ (\%) & PQ$^{th}$ (\%) & PQ$^{st}$ (\%)\\
\midrule\midrule
P 3-7 & P 3-7 & 40.5 & 48.3 & 28.6 \\
P 2-6 & P 3-7 & 40.5 & 48.6 & 28.3 \\
P 2-6 & P 2-6 & {\bf 42.2} & {\bf 49.3} & {\bf 31.4} \\
\bottomrule
\end{tabular}
\caption{
Various results using different FPN levels.
Various results can be obtained by using features from different levels of FPN for the input of Panoptic-Feature generator (P.F.G) and filter sampling module (F.S.M).
The scores are measured on COCO validation set.
}
\vspace{1mm}
\label{Table:FPNlevel}
\end{table}
\begin{table}
\centering 
\begin{tabular}{c|c|cc}
\toprule
Modified & PQ (\%) & Params (M) & FLOPs (10$^{12}$)\\
\midrule
\midrule
& 60.9 & 42.1 & 0.76\\
\checkmark & 63.0 & 42.2 & 0.81\\
\bottomrule
\end{tabular}
\caption{
Effectiveness of modifying ResNet stem module.
The input stem module of ResNet~\cite{ResNet} contains a $7 \times 7$ convolution.
Our model gains huge benefit from simply converting the $7 \times 7$ to three $3 \times 3$~\cite{ResNetStem}.
}
\vspace{-3mm}
\label{Table:Stem}
\end{table}

\paragraph{ResNet Stem Module Modification.}
He~\etal~\cite{ResNetStem} emphasized that classification accuracy of ResNet can be improved by simply modifying the first convolution, which is known as stem module.
By default, the stem module downsizes the input image in half with a $7 \times 7$ convolution.
With the stem module replaced by three $3 \times 3$ convolutions, much finer details can be preserved.
The rich details can lead to substantial gain of performance to the tasks like semantic segmentation which need pixel-level predictions.
Therefore, DeepLabV3+~\cite{DeepLabv3plus} can gain approximate 2\% mIoU increment with the simple modification~\cite{Detectron2}.
Similarly, our model benefits 2.1\% PQ with the modified stem module as shown in~\Tref{Table:Stem}.
Unlike Mask R-CNN, which generates coarse segmentation masks for instances, our model generates fine-grained mask predictions, thus the benefit from the replacement is significant.

\paragraph{Path Integration.}
The integration of the separated branches makes the sharing of computations between things and stuffs become possible.
Though it is clear that our model can benefit huge efficiency from the integration, it is uncertain to expect the gain in terms of performance as the number of parameters decreases.
Therefore, we generated two feature maps from two individual Panoptic-Feature generators, each taking in charge of things and stuffs respectively.
As shown in~\Tref{Table:Unify}, the model with separate feature maps gets a lot heavier than the model with the integrated pathway.
With fewer computations, the integrated model surprisingly overtakes the model with the separated branches by 0.7\% increment of PQ.

\begin{table}
\centering 
{\footnotesize
\begin{tabular}{c|c|ccc}
\toprule
Integrated & PQ (\%) & Speed (ms) & Params (M) & FLOPs (10$^{12}$)\\
\midrule
\midrule
& 62.3 & 202 & 46.6 & 0.96\\
\checkmark & 63.0 & 171 & 42.2 & 0.81\\
\bottomrule
\end{tabular}
}
\caption{
Integrating instance and semantic branches.
The integration lightens both the speed and the computation, while also increasing the performance as implicit interchanges of information between thing and stuff become possible.
}
\vspace{-4mm}
\label{Table:Unify}
\end{table}
\section{Conclusion}
We have introduced SPINet, a model with the novel integration of pathways between the instance and semantic segmentation, which makes efficient yet powerful panoptic segmentation become possible.
With the simplification of the execution flow, SPINet can take the advantages of top-down and bottom-up models, improving even further with proposed auxiliary tasks.
Finally, SPINet shows comparable results on COCO, and delivers the state-of-the-art performance on Cityscapes with a large margin.

{\small
\bibliographystyle{ieee_fullname}

\bibliography{arxiv}
}

\end{document}